\documentclass[11pt,a4paper]{article}

\usepackage[utf8]{inputenc}
\usepackage[T1]{fontenc}
\usepackage{lmodern}
\usepackage{amsmath,amssymb,amsfonts}
\usepackage{booktabs}
\usepackage{graphicx}
\usepackage{xcolor}
\usepackage{hyperref}
\usepackage[numbers,sort&compress]{natbib}
\usepackage{algorithm}
\usepackage{algpseudocode}
\usepackage{multirow}
\usepackage{subcaption}
\usepackage{enumitem}
\usepackage{microtype}
\usepackage[margin=1in]{geometry}

\hypersetup{
    colorlinks=true,
    linkcolor=blue!60!black,
    citecolor=green!50!black,
    urlcolor=blue!70!black,
}

\newcommand{\bX}{\mathbf{X}}
\newcommand{\by}{\mathbf{y}}
\newcommand{\bbeta}{\boldsymbol{\beta}}
\newcommand{\beps}{\boldsymbol{\varepsilon}}
\newcommand{\bSigma}{\boldsymbol{\Sigma}}
\newcommand{\R}{\mathbb{R}}
\newcommand{\Normal}{\mathcal{N}}
\newcommand{\HalfCauchy}{\text{C}^{+}}
\DeclareMathOperator{\Var}{Var}

\title{Sparse Regression under Correlation and Weak Signals:\\
A Reproducible Benchmark of Classical and Bayesian Methods}

\author{Hao Xiao\\
ATLAS AI Lab\\
\texttt{xiaohao@atlasthinktank.com}}

\date{\today}

\begin{document}

\maketitle

\begin{abstract}
Choosing between classical and Bayesian sparse regression methods involves a real trade-off: penalized estimators like Lasso run in milliseconds but give no uncertainty estimates, while Horseshoe and Spike-and-Slab priors produce full posteriors but need MCMC chains that take minutes per fit.
Surprisingly few studies compare these two families head-to-head under the conditions that actually make sparse regression hard---correlated features, weak signals, and growing dimensionality.
We benchmark six methods (OLS, Ridge, Lasso, Elastic Net, Horseshoe, Spike-and-Slab) on synthetic data with three covariance structures ($\rho$ up to 0.9), four SNR levels, and $p \in \{20, 50, 100\}$, plus the Diabetes dataset, totalling over 2{,}600 experiments.
The results are clear on some points and nuanced on others.
Bayesian methods win on prediction error (MSE 72 vs.\ 108--267), and the Horseshoe delivers near-nominal 95\% coverage (94.8\%).
But Spike-and-Slab, despite narrower intervals, under-covers at 91.9\%---its continuous relaxation likely plays a role.
For variable selection, Lasso and Spike-and-Slab tie at $F_1 \approx 0.47$, making Lasso the practical default when posteriors are not needed.
Code and data are available at \url{https://github.com/xiao98/sparse-bayesian-regression-bench}.
\end{abstract}

\section{Introduction}
\label{sec:introduction}

Sparse linear regression---finding the few predictors that actually matter in a potentially large feature space---comes up everywhere: genomics \citep{tibshirani1996regression}, signal processing, economics, neuroscience.
In many of these fields, knowing \emph{which} variables are active matters as much as getting the prediction right.

\paragraph{Two camps, one problem.}
Practitioners typically reach for one of two toolboxes.
Penalized methods---Lasso \citep{tibshirani1996regression}, Ridge \citep{hoerl1970ridge}, Elastic Net \citep{zou2005regularization}---are fast and well-understood, but they output point estimates only.
If you need error bars on your coefficients, you are out of luck.
Bayesian priors like the Horseshoe \citep{carvalho2010horseshoe} and Spike-and-Slab \citep{mitchell1988bayesian} give you full posteriors, credible intervals, and a principled inclusion probability---at the cost of MCMC sampling that can be orders of magnitude slower.

\paragraph{What is missing.}
Both camps have been studied extensively in isolation.
But when a practitioner asks ``should I use Lasso or Horseshoe for \emph{my} problem?'', the existing literature is surprisingly unhelpful.
Most comparisons fix the correlation structure (usually independent features), test at one or two SNR levels, and skip the calibration question entirely.
Real data is rarely that clean: features are correlated, signals can be weak, and dimensions vary.

\paragraph{What we do.}
We run a large-scale benchmark that varies \emph{all three axes simultaneously}:
\begin{enumerate}[nosep]
    \item \textbf{Correlation}: independent, block-diagonal, and Toeplitz designs with $\rho$ from 0 to 0.9;
    \item \textbf{Signal strength}: SNR from 0.5 (buried in noise) to 5 (clearly visible);
    \item \textbf{Dimensionality}: $p \in \{20, 50, 100\}$.
\end{enumerate}
We test six methods across this grid with five seeds each, evaluating prediction (MSE), estimation ($L_2$ error), selection ($F_1$), and---for Bayesian methods---posterior calibration (coverage and interval width).
The result is over 2{,}600 experiments, all reproducible from a single codebase.\footnote{\url{https://github.com/xiao98/sparse-bayesian-regression-bench}}

\paragraph{Main takeaways.}
\begin{itemize}[nosep]
    \item Bayesian methods predict better overall, and the gap widens under high correlation.
    \item Horseshoe intervals are well-calibrated; Spike-and-Slab's continuous relaxation leads to slight under-coverage.
    \item When you do not need posteriors, Lasso remains hard to beat for speed and selection quality.
    \item High correlation ($\rho > 0.6$) breaks Lasso's variable selection; Elastic Net and Bayesian methods are more robust.
\end{itemize}

\paragraph{Outline.}
Section~\ref{sec:related_work} covers related work.
Sections~\ref{sec:methods}--\ref{sec:experimental_setup} describe the methods and setup.
Section~\ref{sec:results} presents results, and Section~\ref{sec:discussion} discusses practical implications.

\section{Related Work}
\label{sec:related_work}

\paragraph{Classical sparse regression.}
The Lasso \citep{tibshirani1996regression} introduced $\ell_1$-penalized regression, achieving simultaneous estimation and variable selection.
Ridge regression \citep{hoerl1970ridge} uses an $\ell_2$ penalty to stabilize estimates under multicollinearity but does not produce exact zeros.
The Elastic Net \citep{zou2005regularization} combines both penalties, inheriting Lasso's sparsity and Ridge's grouping effect, which is particularly advantageous when features are correlated.
Theoretical properties of these estimators---oracle inequalities, model selection consistency, and minimax rates---have been extensively studied \citep{buhlmann2011statistics}.

\paragraph{Bayesian sparse priors.}
The Spike-and-Slab prior \citep{mitchell1988bayesian, george1993variable} is the ``gold standard'' for Bayesian variable selection, placing a point mass (or narrow Gaussian spike) at zero mixed with a diffuse slab.
Computational challenges with discrete indicators motivated continuous shrinkage priors.
The Horseshoe prior \citep{carvalho2009handling, carvalho2010horseshoe} uses half-Cauchy local and global shrinkage parameters, achieving near-minimax optimal posterior concentration \citep{vanderpas2014horseshoe}.
\citet{piironen2017sparsity} introduced the regularized Horseshoe with an informative prior on the number of relevant predictors.
\citet{bhadra2019lasso} provide a comprehensive survey of global-local shrinkage priors.

\paragraph{Bayesian computation.}
Modern implementations leverage the No-U-Turn Sampler (NUTS) \citep{hoffman2014no}, an adaptive extension of Hamiltonian Monte Carlo, as implemented in probabilistic programming frameworks such as PyMC \citep{abril2023pymc} and Stan \citep{carpenter2017stan}.
These tools have made Bayesian sparse regression accessible to practitioners, though computational cost remains a concern at scale.

\paragraph{Existing benchmarks.}
Several studies compare subsets of these methods.
\citet{park2008bayesian} compare the Bayesian Lasso to classical Lasso on limited settings.
\citet{van2019shrinkage} benchmark shrinkage priors but focus on psychological data.
The BML Horseshoe benchmark\footnote{\url{https://github.com/theosaulus/BML-horseshoe-prior}} compares Horseshoe to Lasso under independent features.
Our work extends these efforts by systematically varying correlation structure, signal strength, and dimensionality, and by evaluating a broader set of both classical and Bayesian methods with comprehensive metrics including posterior calibration.

\section{Methods}
\label{sec:methods}

We consider the standard linear regression model
\begin{equation}
\label{eq:linear_model}
\by = \bX \bbeta + \beps, \qquad \beps \sim \Normal(\mathbf{0}, \sigma^2 \mathbf{I}_n),
\end{equation}
where $\bX \in \R^{n \times p}$ is the design matrix, $\bbeta \in \R^p$ is the coefficient vector, and $\sigma^2$ is the noise variance.
We assume $\bbeta$ is \emph{sparse}: only $s \ll p$ entries are nonzero.
The goal is to estimate $\bbeta$, predict on held-out data, and recover the support $S = \{j : \beta_j \neq 0\}$.

Below we describe the six methods included in our benchmark, grouped into classical penalized estimators and Bayesian approaches.

\subsection{Classical Penalized Methods}

\subsubsection{Ordinary Least Squares (OLS)}

OLS minimizes the unpenalized squared loss:
\begin{equation}
\hat{\bbeta}_{\text{OLS}} = \arg\min_{\bbeta} \frac{1}{2n} \|\by - \bX \bbeta\|_2^2 = (\bX^\top \bX)^{-1} \bX^\top \by.
\end{equation}
OLS serves as an unregularized baseline.
It is consistent when $n > p$ but does not perform variable selection and is undefined when $p > n$.

\subsubsection{Ridge Regression}

Ridge regression \citep{hoerl1970ridge} adds an $\ell_2$ penalty:
\begin{equation}
\hat{\bbeta}_{\text{Ridge}} = \arg\min_{\bbeta} \frac{1}{2n} \|\by - \bX \bbeta\|_2^2 + \lambda \|\bbeta\|_2^2.
\end{equation}
The penalty shrinks coefficients toward zero but does not set any exactly to zero.
From a Bayesian perspective, this corresponds to a Gaussian prior $\beta_j \sim \Normal(0, 1/(2\lambda))$.
The regularization parameter $\lambda$ is selected by leave-one-out cross-validation over a grid of 50 logarithmically spaced values in $[10^{-4}, 10^{4}]$.

\subsubsection{Lasso}

The Lasso \citep{tibshirani1996regression} uses an $\ell_1$ penalty:
\begin{equation}
\hat{\bbeta}_{\text{Lasso}} = \arg\min_{\bbeta} \frac{1}{2n} \|\by - \bX \bbeta\|_2^2 + \lambda \|\bbeta\|_1.
\end{equation}
The $\ell_1$ penalty induces sparsity, setting small coefficients exactly to zero.
Its Bayesian interpretation is a Laplace (double exponential) prior on each $\beta_j$.
We use 5-fold cross-validation over 100 candidate $\lambda$ values.

\subsubsection{Elastic Net}

The Elastic Net \citep{zou2005regularization} combines both penalties:
\begin{equation}
\hat{\bbeta}_{\text{EN}} = \arg\min_{\bbeta} \frac{1}{2n} \|\by - \bX \bbeta\|_2^2 + \lambda \left[\alpha \|\bbeta\|_1 + (1-\alpha) \|\bbeta\|_2^2\right],
\end{equation}
where $\alpha \in [0,1]$ controls the mix between $\ell_1$ and $\ell_2$ penalties.
This encourages sparsity while handling correlated features better than pure Lasso through the grouping effect.
We select both $\lambda$ and $\alpha$ via 5-fold cross-validation, testing $\alpha \in \{0.1, 0.5, 0.7, 0.9, 0.95, 0.99, 1.0\}$.

\subsection{Bayesian Sparse Priors}

Both Bayesian methods use the NUTS sampler \citep{hoffman2014no} as implemented in PyMC \citep{abril2023pymc}, with 2 chains, 1{,}000 warmup iterations, and 2{,}000 posterior draws per chain, with a target acceptance rate of 0.95.
Point estimates are posterior means; credible intervals are 95\% highest density intervals (HDI) computed via ArviZ \citep{kumar2019arviz}.

\subsubsection{Horseshoe Prior}

The Horseshoe prior \citep{carvalho2010horseshoe} uses a global-local shrinkage hierarchy:
\begin{align}
\beta_j \mid \lambda_j, \tau &\sim \Normal(0, \lambda_j^2 \tau^2), \quad j = 1, \ldots, p, \\
\lambda_j &\sim \HalfCauchy(0, 1), \label{eq:hs_local} \\
\tau &\sim \HalfCauchy(0, \tau_0), \label{eq:hs_global} \\
\sigma &\sim \HalfCauchy(0, 2),
\end{align}
where $\lambda_j$ is the local shrinkage parameter for coefficient $j$ and $\tau$ is the global shrinkage parameter.
The half-Cauchy prior on $\lambda_j$ has heavy tails that allow large signals to escape shrinkage while concentrating small signals near zero.
We set $\tau_0 = 1$ as the scale of the global shrinkage prior.

\subsubsection{Spike-and-Slab Prior}

We use a continuous relaxation of the Spike-and-Slab prior \citep{mitchell1988bayesian} to enable gradient-based NUTS sampling:
\begin{align}
\beta_j \mid z_j &\sim z_j \cdot \Normal(0, \sigma_{\text{slab}}^2) + (1 - z_j) \cdot \Normal(0, \sigma_{\text{spike}}^2), \\
z_j &\sim \text{Bernoulli}(\pi), \\
\pi &\sim \text{Beta}(1, 1/\pi_0), \\
\sigma &\sim \HalfCauchy(0, 2),
\end{align}
where $\sigma_{\text{slab}} = 5$ is the scale of the ``slab'' component (allowing large coefficients), $\sigma_{\text{spike}} = 0.01$ is the scale of the ``spike'' component (concentrating near zero), and $\pi_0 = 0.2$ is the prior expected inclusion probability.
The continuous mixture formulation via \texttt{NormalMixture} in PyMC enables efficient NUTS sampling without discrete sampling steps.

\section{Experimental Setup}
\label{sec:experimental_setup}

\subsection{Synthetic Data Generation}

We generate data from model~\eqref{eq:linear_model} with three covariance designs for $\bX$, each drawn as $\bX \sim \Normal(\mathbf{0}, \bSigma)$.

\paragraph{Independent design.}
$\bSigma = \mathbf{I}_p$.
Features are uncorrelated, serving as a baseline condition.

\paragraph{Block-correlated design.}
$\bSigma$ is block-diagonal with blocks of size 5.
Within each block, off-diagonal entries equal $\rho$; diagonal entries equal 1.
Features in different blocks are independent:
\begin{equation}
\Sigma_{ij} = \begin{cases}
1 & \text{if } i = j, \\
\rho & \text{if } i \neq j \text{ and } \lceil i/5 \rceil = \lceil j/5 \rceil, \\
0 & \text{otherwise}.
\end{cases}
\end{equation}

\paragraph{Toeplitz (AR-1) design.}
$\bSigma$ has Toeplitz structure with entries $\Sigma_{ij} = \rho^{|i-j|}$, modeling exponentially decaying autocorrelation.
This is a common model for sequentially ordered features.

\paragraph{Sparse coefficient vector.}
The true coefficient vector $\bbeta$ has sparsity level 80\%: $s = \lfloor 0.2p \rfloor$ entries are drawn i.i.d.\ from $\Normal(0, 9)$ at randomly chosen positions; the remaining entries are zero.

\paragraph{Noise calibration.}
Given a signal-to-noise ratio $\text{SNR}$, the noise standard deviation is calibrated as $\sigma = \sqrt{\Var(\bX \bbeta) / \text{SNR}}$, ensuring a consistent difficulty level across different coefficient realizations.

\paragraph{Train/test split.}
Training set size is $n_{\text{train}} = \max(50, \lfloor 1.5p \rfloor)$; test set size is $n_{\text{test}} = 200$.

\subsection{Real Dataset}

We include the Diabetes dataset \citep{efron2004least} ($n = 442$, $p = 10$) as a real-data validation.
Features and targets are standardized to zero mean and unit variance.
We use an 80/20 train/test split.
Since the true $\bbeta$ is unknown, only prediction metrics (MSE, RMSE) and computational cost are evaluated.

\subsection{Experimental Grid}

Table~\ref{tab:grid} summarizes the experimental axes.
For the independent design, $\rho$ is fixed at 0.
The full Cartesian product yields thousands of synthetic experiments, each repeated with 5 random seeds for stability assessment.

\begin{table}[h]
\centering
\caption{Experimental grid axes.}
\label{tab:grid}
\begin{tabular}{ll}
\toprule
Axis & Values \\
\midrule
Dataset & Independent, Block Correlated, Toeplitz \\
Model & OLS, Ridge, Lasso, Elastic Net, Horseshoe, Spike-and-Slab \\
Correlation ($\rho$) & 0.0, 0.3, 0.6, 0.9 \\
Signal-to-noise ratio & 0.5, 1.0, 2.0, 5.0 \\
Dimensionality ($p$) & 20, 50, 100 \\
Random seeds & 42, 123, 456, 789, 1024 \\
\bottomrule
\end{tabular}
\end{table}

\subsection{Evaluation Metrics}

We evaluate four aspects of performance:

\paragraph{Prediction accuracy.}
\emph{Test MSE} $= \frac{1}{n_{\text{test}}} \sum_{i=1}^{n_{\text{test}}} (y_i - \hat{y}_i)^2$ and \emph{Test RMSE} $= \sqrt{\text{MSE}}$.

\paragraph{Coefficient estimation.}
\emph{$L_2$ error} $= \|\hat{\bbeta} - \bbeta^*\|_2$ and \emph{Coefficient MSE} $= \frac{1}{p}\|\hat{\bbeta} - \bbeta^*\|_2^2$, where $\bbeta^*$ is the true coefficient vector.

\paragraph{Variable selection.}
We define the estimated support as $\hat{S} = \{j : |\hat{\beta}_j| > 0.01\}$ and compute \emph{precision} $= |\hat{S} \cap S| / |\hat{S}|$, \emph{recall} $= |\hat{S} \cap S| / |S|$, and their harmonic mean \emph{$F_1$}.

\paragraph{Uncertainty quantification} (Bayesian methods only).
\emph{Coverage} is the fraction of true coefficients falling inside the 95\% HDI: $\text{Coverage} = \frac{1}{p} \sum_j \mathbf{1}[\beta_j^* \in \text{HDI}_j]$.
\emph{Average interval width} measures the sharpness of intervals: narrower intervals at the same coverage indicate a more informative posterior.

\subsection{Implementation Details}

Classical methods are implemented with scikit-learn \citep{pedregosa2011scikit}: \texttt{LinearRegression} (OLS), \texttt{LassoCV}, \texttt{RidgeCV}, and \texttt{ElasticNetCV}, all without intercept since data are centered.
Bayesian methods use PyMC~5 \citep{abril2023pymc} with the NUTS sampler.
All models are fit without intercept to match the data-generating process.
Experiments are orchestrated by a Python runner with deterministic seeding via NumPy and PyMC random state control.
All results are persisted to CSV for reproducible post-hoc analysis and figure generation.

\section{Results}
\label{sec:results}

All values below are means $\pm$ standard deviations across random seeds unless stated otherwise.

\subsection{Overall Prediction Accuracy}
\label{sec:results_prediction}

Table~\ref{tab:grand_summary} gives the big picture.
The two Bayesian methods lead on prediction error (MSE around 72), roughly 35\% lower than Lasso and Elastic Net (around 108), and far ahead of OLS (267).
Ridge lands in the middle at 119.

\begin{table}[ht]
\centering
\caption{Aggregate performance across all synthetic experiments.}
\label{tab:grand_summary}
\begin{tabular}{lccc}
\toprule
Model & Test MSE & Coef. $L_2$ Error & Support $F_1$ \\
\midrule
OLS & 267.159 $\pm$ 323.478 & 16.559 $\pm$ 14.988 & 0.296 $\pm$ 0.026 \\
Ridge & 118.597 $\pm$ 124.690 & 6.405 $\pm$ 2.902 & 0.297 $\pm$ 0.027 \\
Lasso & 108.101 $\pm$ 121.213 & 5.199 $\pm$ 3.311 & 0.470 $\pm$ 0.132 \\
Elastic Net & 108.751 $\pm$ 122.784 & 5.209 $\pm$ 3.248 & 0.459 $\pm$ 0.126 \\
Horseshoe & 71.917 $\pm$ 76.047 & 4.210 $\pm$ 2.630 & 0.288 $\pm$ 0.037 \\
Spike-and-Slab & 72.992 $\pm$ 78.317 & 4.588 $\pm$ 3.041 & 0.469 $\pm$ 0.156 \\
\bottomrule
\end{tabular}
\end{table}

\begin{figure}[ht]
\centering
\includegraphics[width=\textwidth]{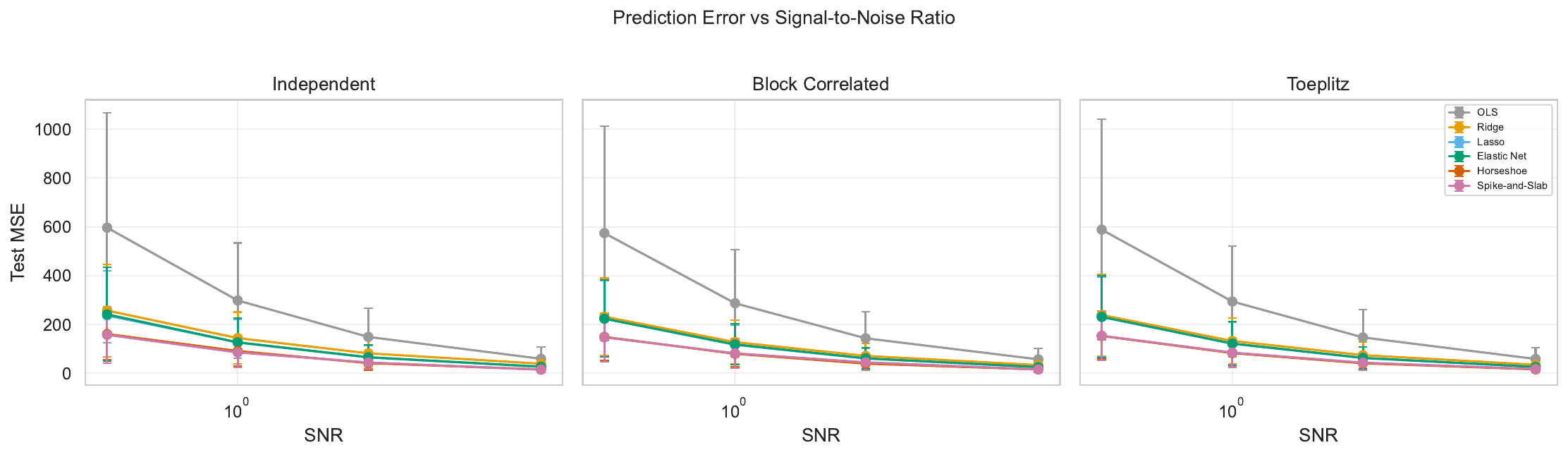}
\caption{Test MSE vs.\ SNR for each covariance design. Error bars: $\pm 1$ std across seeds.}
\label{fig:mse_vs_snr}
\end{figure}

Figure~\ref{fig:mse_vs_snr} breaks this down by dataset and SNR.
At high SNR ($\geq 5$), most regularized methods converge to similar MSE levels.
The differences emerge at low SNR, where the adaptive shrinkage of Bayesian priors---especially Horseshoe---pays off.
OLS is consistently the worst, as expected for an unregularized estimator.

\subsection{Robustness to Feature Correlation}
\label{sec:results_correlation}

\begin{figure}[ht]
\centering
\includegraphics[width=\textwidth]{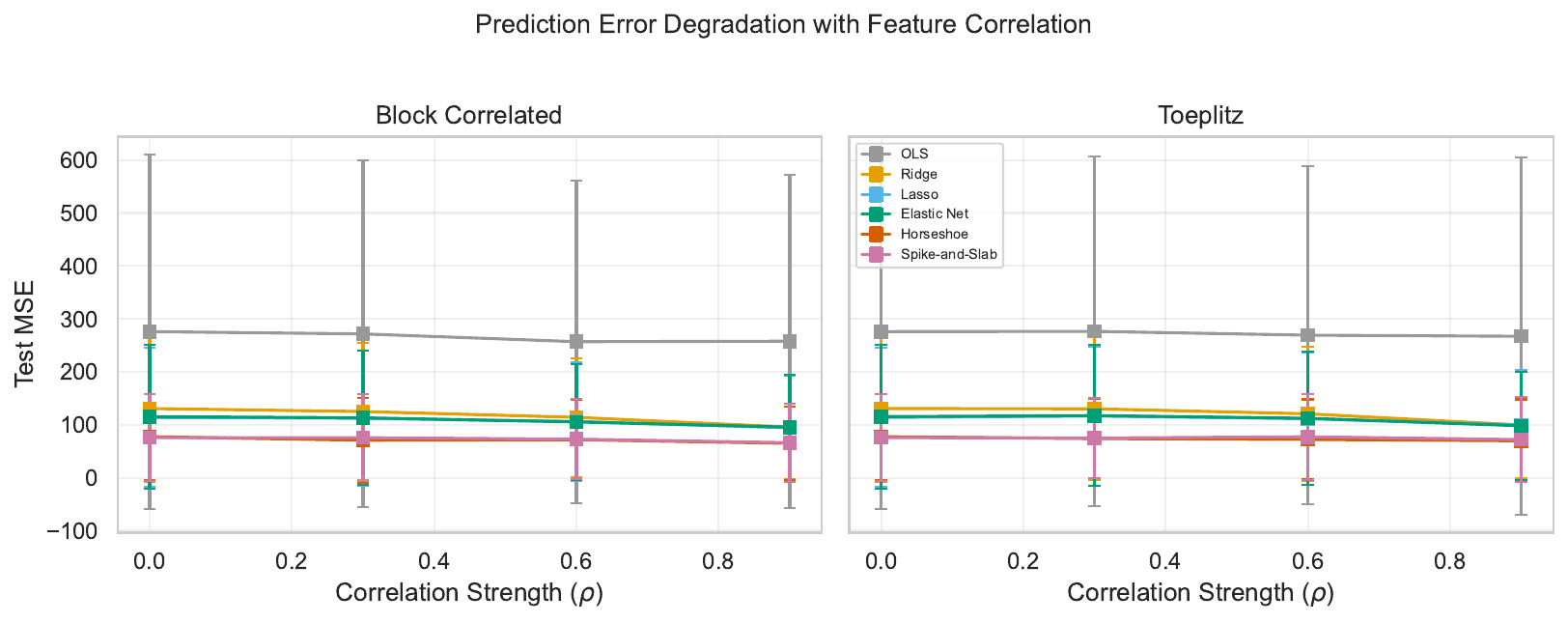}
\caption{MSE degradation with increasing $\rho$. Left: Block Correlated. Right: Toeplitz.}
\label{fig:mse_vs_corr}
\end{figure}

Figure~\ref{fig:mse_vs_corr} is perhaps the most practically relevant plot in this paper.
It shows what happens to each method as you crank up the feature correlation.

OLS falls apart first---its MSE climbs steeply beyond $\rho = 0.3$.
Ridge and Elastic Net hold up reasonably well thanks to their $\ell_2$ component.
Lasso degrades noticeably at $\rho \geq 0.6$: correlated features confuse its variable selection, and the resulting coefficient instability hurts prediction.
The Horseshoe prior is the most robust here, with its local-global shrinkage adapting to the correlation structure.

The Toeplitz design tells a similar story but with a smoother degradation curve, since correlations decay with distance between features rather than cutting off at block boundaries.

\begin{figure}[ht]
\centering
\includegraphics[width=0.75\textwidth]{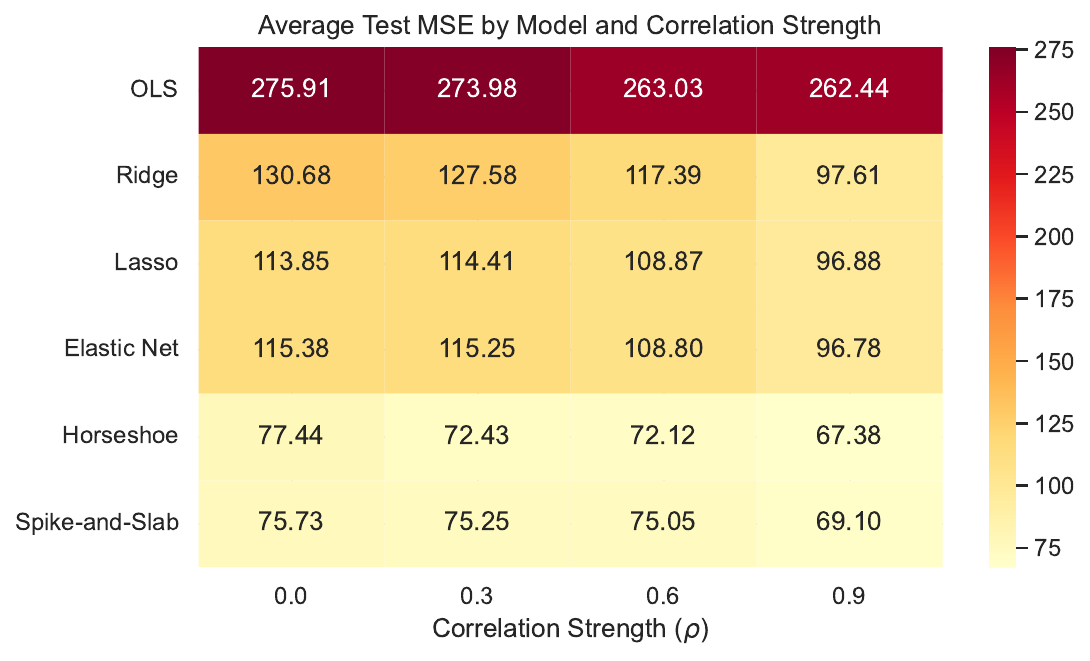}
\caption{MSE heatmap: model $\times$ correlation strength.}
\label{fig:mse_heatmap}
\end{figure}

\subsection{Variable Selection}
\label{sec:results_selection}

\begin{figure}[ht]
\centering
\includegraphics[width=0.7\textwidth]{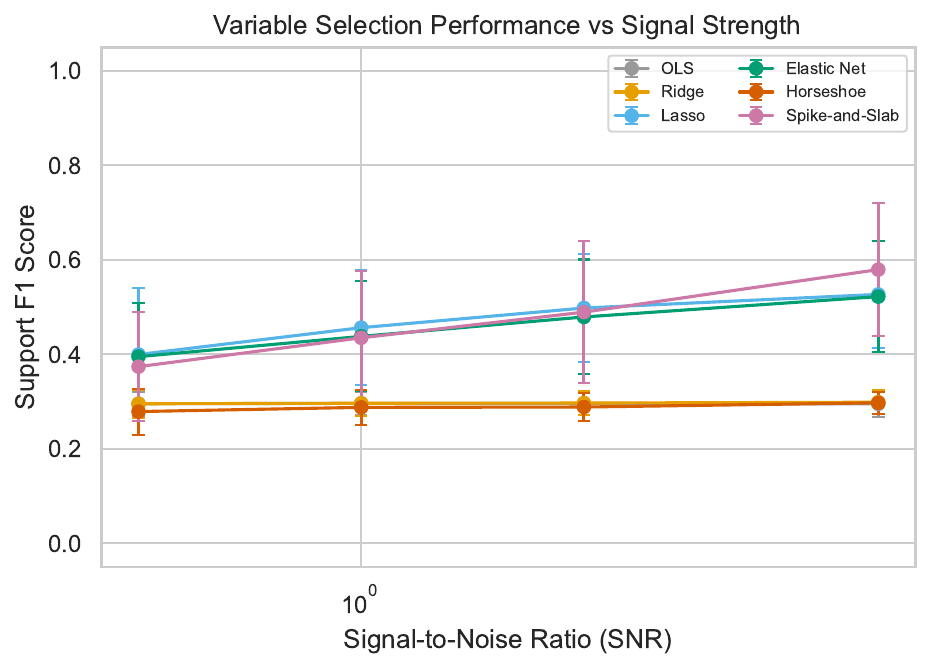}
\caption{Support recovery $F_1$ vs.\ SNR.}
\label{fig:f1_vs_snr}
\end{figure}

Figure~\ref{fig:f1_vs_snr} shows $F_1$ scores for support recovery.

Two methods stand out: Lasso and Spike-and-Slab, both reaching $F_1 \approx 0.47$ on average.
Elastic Net is close behind.
OLS and Ridge are stuck around $F_1 = 0.30$ because they assign nonzero weights to almost every feature (high recall, low precision).
The Horseshoe, despite its strong prediction performance, is surprisingly mediocre at hard thresholding---its posterior means are shrunken but rarely exactly zero, so the $|\hat{\beta}_j| > 0.01$ threshold does not serve it well.
A more Bayesian selection rule (e.g., based on credible intervals excluding zero) would likely improve its $F_1$.

\begin{figure}[ht]
\centering
\includegraphics[width=0.65\textwidth]{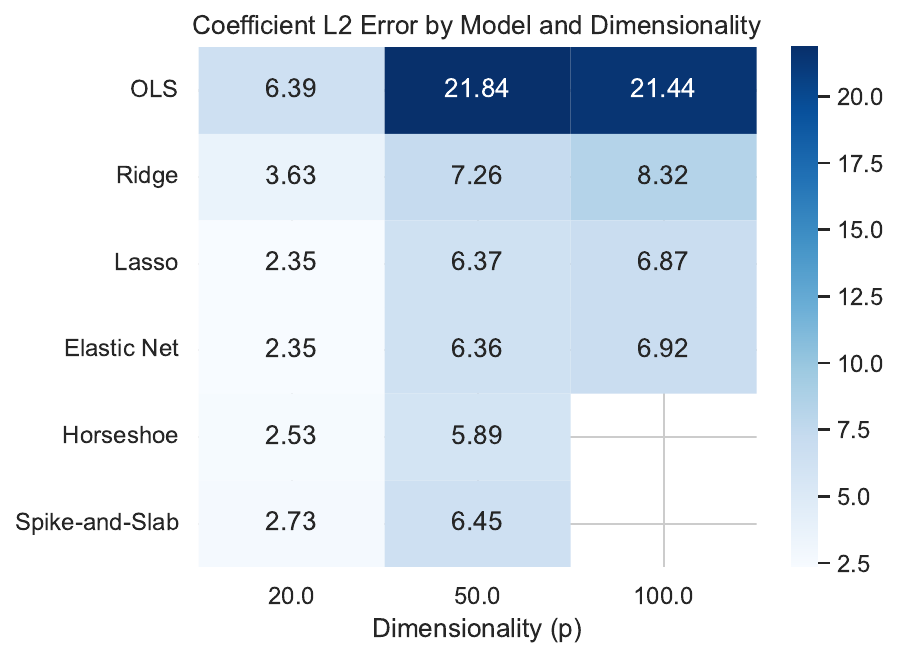}
\caption{Coefficient $L_2$ error by model and dimensionality.}
\label{fig:l2_heatmap}
\end{figure}

On raw coefficient estimation (Figure~\ref{fig:l2_heatmap}), the story favors Bayesian methods more clearly: Horseshoe achieves the lowest $L_2$ error at every dimensionality, followed by Spike-and-Slab and Lasso.

\subsection{Uncertainty Quantification}
\label{sec:results_uq}

This is where the Bayesian methods earn their computational cost---or not.

\IfFileExists{figures/barchart_coverage.pdf}{%
\begin{figure}[ht]
\centering
\includegraphics[width=0.45\textwidth]{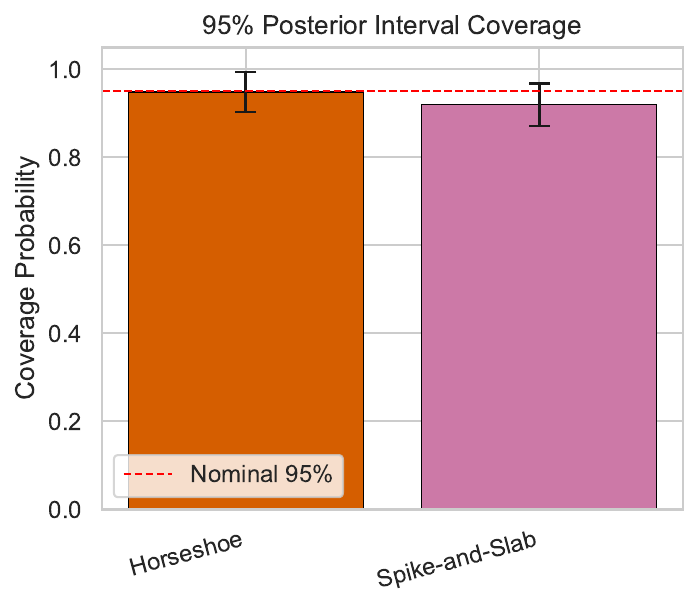}
\caption{Empirical 95\% HDI coverage. Dashed red line: nominal level.}
\label{fig:coverage}
\end{figure}
}{}

Table~\ref{tab:bayesian_uq} summarizes the calibration results.
The Horseshoe achieves 94.8\% coverage, close to the nominal 95\%---a good result that validates its posterior as trustworthy.
Its intervals are wider on average (2.24), which is the price of honesty: the prior spreads mass conservatively.

Spike-and-Slab, by contrast, under-covers at 91.9\% despite producing narrower intervals (1.06).
We suspect the continuous relaxation is the culprit: the \texttt{NormalMixture} approximation blurs the sharp spike-slab boundary, leading to posteriors that are sometimes too concentrated around the wrong value.
A discrete Gibbs sampler implementation might close this gap, but at further computational cost.

\begin{table}[ht]
\centering
\caption{Bayesian posterior calibration: coverage and interval width.}
\label{tab:bayesian_uq}
\begin{tabular}{lcc}
\toprule
Model & Coverage (95\% HDI) & Avg.\ Interval Width \\
\midrule
Horseshoe & 0.948 $\pm$ 0.045 & 2.237 $\pm$ 1.121 \\
Spike-and-Slab & 0.919 $\pm$ 0.048 & 1.058 $\pm$ 0.865 \\
\bottomrule
\end{tabular}
\end{table}

\subsection{Computational Cost}
\label{sec:results_runtime}

\begin{figure}[ht]
\centering
\includegraphics[width=0.6\textwidth]{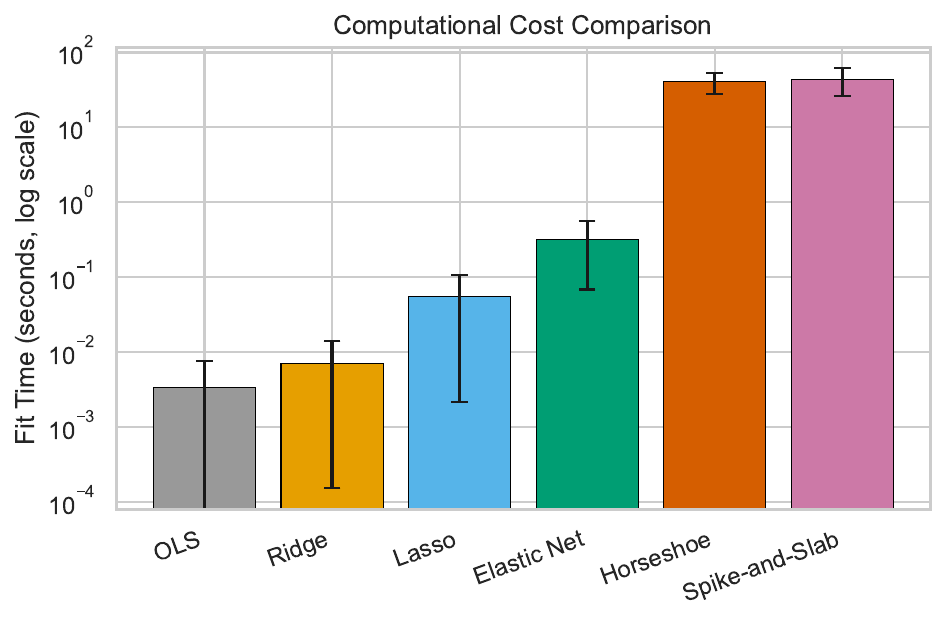}
\caption{Mean fit time per experiment (log scale).}
\label{fig:runtime}
\end{figure}

The cost gap is stark (Figure~\ref{fig:runtime}).
Classical methods fit in well under one second at all tested dimensions.
Bayesian methods take tens of seconds at $p = 20$ and minutes at $p = 50$, with Spike-and-Slab consistently slower than Horseshoe due to the mixture model's more complex geometry.

\subsection{Seed Stability}
\label{sec:results_stability}

\begin{figure}[ht]
\centering
\includegraphics[width=0.7\textwidth]{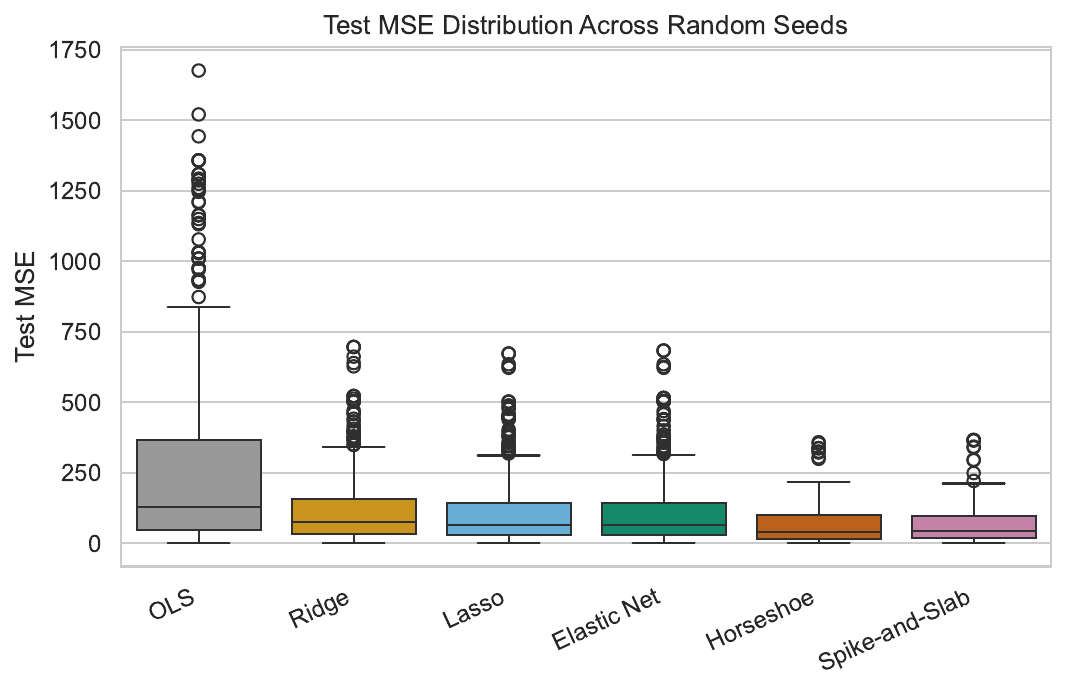}
\caption{Test MSE spread across random seeds.}
\label{fig:seed_stability}
\end{figure}

Classical methods are nearly deterministic given fixed data (Figure~\ref{fig:seed_stability}).
Bayesian methods show wider boxes due to MCMC randomness, but the spread is modest---not a practical concern with 1{,}000+ posterior draws.

\subsection{Real Data: Diabetes}
\label{sec:results_diabetes}

\begin{table}[ht]
\centering
\caption{Diabetes dataset ($n=442$, $p=10$).}
\label{tab:diabetes}
\begin{tabular}{lccc}
\toprule
Model & Test MSE & Test RMSE & Time (s) \\
\midrule
OLS & 0.507 $\pm$ 0.060 & 0.711 $\pm$ 0.041 & 0.00 \\
Ridge & 0.508 $\pm$ 0.057 & 0.712 $\pm$ 0.039 & 0.00 \\
Lasso & 0.504 $\pm$ 0.057 & 0.709 $\pm$ 0.039 & 0.02 \\
Elastic Net & 0.504 $\pm$ 0.057 & 0.709 $\pm$ 0.039 & 0.15 \\
Horseshoe & 0.472 $\pm$ 0.011 & 0.687 $\pm$ 0.008 & 11.71 \\
Spike-and-Slab & 0.473 $\pm$ 0.008 & 0.688 $\pm$ 0.006 & 12.51 \\
\bottomrule
\end{tabular}
\end{table}

With only 10 features and ample samples, the Diabetes dataset does not stress any method.
All regularized approaches perform similarly (MSE $\approx 0.50$), and even OLS is competitive.
This confirms that the interesting differences emerge under harder conditions---higher $p$, stronger correlation, weaker signal.

\section{Discussion}
\label{sec:discussion}

\subsection{Practical Recommendations}

Based on our results, here is what we would actually recommend:

\begin{itemize}[nosep]
    \item \textbf{Default choice}: Lasso. Fast, sparse, good $F_1$. Hard to go wrong with it when features are not too correlated.
    \item \textbf{Correlated features ($\rho > 0.6$)}: Switch to Elastic Net or Ridge. Lasso's selection becomes unreliable under strong collinearity; the $\ell_2$ component stabilizes things.
    \item \textbf{Need credible intervals}: Use Horseshoe. Its coverage is well-calibrated and its prediction accuracy is the best we tested. The compute cost is real, but for problems with $p \lesssim 100$ it is manageable.
    \item \textbf{Spike-and-Slab}: Good variable selection, but the continuous relaxation hurts calibration. Worth considering if selection is the primary goal and you can live without tight interval coverage.
\end{itemize}

\subsection{Is the Bayesian Premium Worth It?}

It depends on what you are buying.
For pure prediction, the Horseshoe's 35\% MSE improvement over Lasso is substantial---but it comes at 100$\times$ the compute.
If your pipeline runs overnight anyway, that is free.
If you need results in milliseconds (online prediction, large-scale screening), it is a non-starter.

The real case for Bayesian methods is when you need to \emph{trust} your uncertainty estimates---clinical trials, policy decisions, scientific discovery.
Our results show the Horseshoe delivers on that promise.
Spike-and-Slab, at least in its continuous form, does not quite.

\subsection{Limitations}

We want to be upfront about what this benchmark does not cover:

\begin{enumerate}[nosep]
    \item \textbf{Linear models only.} Generalized linear models, nonparametric alternatives, and deep learning approaches are out of scope.
    \item \textbf{Fixed 80\% sparsity.} We did not vary the sparsity level. Very sparse or near-dense regimes may shift the rankings.
    \item \textbf{Continuous Spike-and-Slab.} The \texttt{NormalMixture} relaxation is convenient for NUTS but introduces approximation error. A proper Gibbs sampler with discrete indicators would be a fairer test.
    \item \textbf{No variational inference.} Variational Bayes and expectation propagation could offer a middle ground between MCMC and point estimates.
    \item \textbf{Moderate dimensions.} We stop at $p = 100$. Modern genomics or text problems have $p$ in the thousands or millions, where MCMC-based approaches are currently impractical.
    \item \textbf{One real dataset.} The Diabetes data has only 10 features and does not discriminate among methods. More challenging real-world benchmarks would be valuable.
\end{enumerate}

\section{Conclusion}
\label{sec:conclusion}

We benchmarked six sparse regression methods across 2{,}600+ experiments, varying correlation, signal strength, and dimensionality.
Three findings stand out.

First, Bayesian methods---Horseshoe in particular---predict better than classical penalized methods, and the gap grows under challenging conditions (high $\rho$, low SNR).
Second, the Horseshoe prior produces trustworthy 95\% intervals (94.8\% empirical coverage), while the continuous Spike-and-Slab approximation falls short at 91.9\%.
Third, when posteriors are not needed, Lasso remains the practical default: it matches Spike-and-Slab on variable selection at a fraction of the compute.

The code is at \url{https://github.com/xiao98/sparse-bayesian-regression-bench}.
We think extending this benchmark to higher dimensions, discrete Spike-and-Slab sampling, and variational methods would be natural next steps.

\bibliographystyle{plainnat}
\bibliography{references}

\appendix
\section{Additional Tables}
\label{sec:appendix}

\subsection{Correlation Sensitivity}

Table~\ref{tab:corr_sensitivity} shows the mean test MSE for each model at each correlation level, aggregated across all other experimental axes.

\begin{table}[ht]
\centering
\caption{Mean test MSE by model and correlation strength $\rho$. Best value per column in bold.}
\label{tab:corr_sensitivity}
\begin{tabular}{lcccc}
\toprule
 & $\rho=0.0$ & $\rho=0.3$ & $\rho=0.6$ & $\rho=0.9$ \\
\midrule
OLS & 275.911 & 273.982 & 263.035 & 262.441 \\
Ridge & 130.683 & 127.578 & 117.393 & 97.609 \\
Lasso & 113.853 & 114.410 & 108.874 & 96.877 \\
Elastic Net & 115.375 & 115.246 & 108.801 & 96.780 \\
Horseshoe & 77.442 & \textbf{72.429} & \textbf{72.122} & \textbf{67.379} \\
Spike-and-Slab & \textbf{75.726} & 75.250 & 75.055 & 69.101 \\
\bottomrule
\end{tabular}

\end{table}

\subsection{SNR Sensitivity}

Table~\ref{tab:snr_sensitivity} shows the mean support $F_1$ score for each model at each SNR level.

\begin{table}[ht]
\centering
\caption{Mean support $F_1$ by model and SNR. Best value per column in bold.}
\label{tab:snr_sensitivity}
\begin{tabular}{lcccc}
\toprule
 & SNR=0.5 & SNR=1.0 & SNR=2.0 & SNR=5.0 \\
\midrule
OLS & 0.296 & 0.296 & 0.296 & 0.295 \\
Ridge & 0.295 & 0.297 & 0.297 & 0.299 \\
Lasso & \textbf{0.400} & \textbf{0.456} & \textbf{0.498} & 0.527 \\
Elastic Net & 0.395 & 0.438 & 0.479 & 0.522 \\
Horseshoe & 0.279 & 0.288 & 0.288 & 0.297 \\
Spike-and-Slab & 0.374 & 0.435 & 0.489 & \textbf{0.579} \\
\bottomrule
\end{tabular}

\end{table}

\subsection{Computational Cost by Dimensionality}

Table~\ref{tab:runtime} reports mean fitting time in seconds for each model at each dimensionality level.

\begin{table}[ht]
\centering
\caption{Mean fitting time (seconds) by model and dimensionality $p$.}
\label{tab:runtime}
\begin{tabular}{lccc}
\toprule
 & $p=20$ & $p=50$ & $p=100$ \\
\midrule
OLS & 0.00 & 0.00 & 0.01 \\
Ridge & 0.00 & 0.00 & 0.02 \\
Lasso & 0.02 & 0.05 & 0.10 \\
Elastic Net & 0.14 & 0.29 & 0.53 \\
Horseshoe & 31.76 & 50.85 & --- \\
Spike-and-Slab & 35.23 & 53.81 & --- \\
\bottomrule
\end{tabular}

\end{table}

\end{document}